\documentclass[letterpaper]{article} % DO NOT CHANGE THIS
\usepackage{aaai2026}  % DO NOT CHANGE THIS
\usepackage{times}  % DO NOT CHANGE THIS
\usepackage{helvet}  % DO NOT CHANGE THIS
\usepackage{courier}  % DO NOT CHANGE THIS
\usepackage[hyphens]{url}  % DO NOT CHANGE THIS
\usepackage{graphicx} % DO NOT CHANGE THIS
\def\UrlFont{\rm}  % DO NOT CHANGE THIS
\usepackage{natbib}  % DO NOT CHANGE THIS AND DO NOT ADD ANY OPTIONS TO IT
\usepackage{caption} % DO NOT CHANGE THIS AND DO NOT ADD ANY OPTIONS TO IT
\usepackage{algorithm}

\usepackage[mathscr]{euscript}
\usepackage[utf8]{inputenc}
\usepackage{amsthm}
\usepackage{amssymb}
\usepackage[noend]{algpseudocode}
\usepackage{stmaryrd}
\usepackage{mathtools}
\newtheorem{definition}{Definition}

\newtheorem{theorem}{Theorem}
\newtheorem{corollary}{Corollary}

\newcommand{\LFP}{\ensuremath{\textup{LFP}}}
\newcommand{\BFP}{\ensuremath{\textup{BFP}}}
\newcommand{\LFPsim}{\ensuremath{\textup{LFP-sim}}}
\newcommand{\lfp}{\ensuremath{\textup{\textbf{lfp}}}}
\newcommand{\FO}{\ensuremath{\textup{FO}}}
\newcommand{\struct}{\ensuremath{\mathcal{S}}}

\newcommand{\prog}{\ensuremath{\mathscr{P}}}
\newcommand{\axquery}{\ensuremath{\mathscr{Q}}}
\newcommand{\semposaxioms}{\ensuremath{\textup{AP}_0}}
\newcommand{\genaxioms}{\ensuremath{\textup{AP}}}

\newcommand{\pedge}{\ensuremath{\pred{E}}}
\newcommand{\ppath}{\ensuremath{\pred{path}}}
\newcommand{\pacyclic}{\ensuremath{\pred{acyclic}}}

\newcommand{\leqij}[2]{\ensuremath{\preceq^{#1,#2}}}
\newcommand{\leij}[2]{\ensuremath{\prec^{#1,#2}}}
\newcommand{\nleqij}[2]{\ensuremath{\npreceq^{#1,#2}}}
\newcommand{\nleij}[2]{\ensuremath{\nprec^{#1,#2}}}
\newcommand{\triij}[2]{\ensuremath{\triangleleft^{#1,#2}}}
\newcommand{\stage}[2]{\ensuremath{|#1|_{P_#2}^\struct}}

\newcommand{\leqi}[2]{\ensuremath{{\preceq^{#1}}#2}}
\newcommand{\lei}[2]{\ensuremath{{\prec^{#1}}#2}}
\newcommand{\nleqi}[2]{\ensuremath{{\npreceq^{#1}}#2}}
\newcommand{\nlei}[2]{\ensuremath{{\nprec^{#1}}#2}}

\newcommand{\pred}[1]{\ensuremath{\textit{#1}}}

\newcommand{\pddlaxiom}[2]{\ensuremath{#1\leftarrow{} #2}}

\newcommand{\inlinecite}[1]{\citeauthor{#1} (\citeyear{#1})}

\title{PDDL Axioms Are Equivalent to Least Fixed Point Logic (Extended Version)}
\author {
  Claudia Grundke,
  Gabriele Röger
}
\affiliations {
  University of Basel \\
  Basel, Switzerland \\
  \{claudia.grundke, gabriele.roeger\}@unibas.ch
}

\nocopyright
\begin{document}

\maketitle

\begin{abstract}
  Axioms are a feature of the Planning Domain Definition Language PDDL that can
  be considered as a generalization of database query languages such as
  Datalog.
  The PDDL standard restricts negative occurrences of predicates in axiom
  bodies to predicates that are directly set by actions and not derived by
  axioms. In the literature, authors often deviate from this limitation and
  only require that the set of axioms is stratifiable.
  We show that both variants can express exactly the same queries as least fixed
  point logic. They are thus strictly more expressive than stratified Datalog,
  which aligns with another restriction on axioms occasionally considered in
  the planning literature.
  Complementing this theoretical analysis, we also present a compilation that
  eliminates negative occurrences of derived predicates from PDDL axioms.
\end{abstract}

This is an extended version of an ICAPS 2026 paper
\cite{grundke-roeger-icaps2026-to-appear}. The main difference is the full proof of
Theorem \ref{thm:stagelemma}.

\section{Introduction}

The aim in classical automated planning is to reach a desirable world state from the
current world state by applying a suitable sequence of actions.
World states are represented as sets of ground atoms
that are true in the state. The set of all ground atoms of the planning task is
defined by a finite set of predicates, each with an associated arity, and the finite
set of constants (or objects) of the task. Ground atoms that do not occur
in the set representation of a state $s$ are false in $s$.

The predicates are partitioned into \emph{basic} and \emph{derived}
predicates. The actions only directly affect the truth of the basic
predicates, whereas the interpretation of the derived predicates is determined
from the interpretation of the basic predicates by means of a logic program,
consisting of so-called axioms. An axiom has the form $\pddlaxiom{P(\vec
x)}{\varphi(\vec x)}$ and expresses that the \emph{head} $P(\vec x)$ is true if
the \emph{body} $\varphi(\vec x)$ is true.

Consider as an example a basic predicate $\pedge$ for an edge relation and
a derived predicate $\ppath$. The axiom 
\begin{equation}
\pddlaxiom{\ppath(x,y)}{\pedge(x,y)\lor \exists z(\pedge(x,z)\land \ppath(z,y))}
\label{eq:path}
\end{equation}
expresses that there is a path from $x$ to $y$ if there is a direct edge from $x$ to
$y$, or if there is an edge from $x$ to a successor $z$ from which there is a path to
$y$. Axioms are evaluated by initializing all derived atoms as false and
successively making them true based on the axioms until a fixed point is
reached. With this example axiom we would therefore interpret $\ppath$ as the
transitive closure of the edge relation $\pedge$.

Capturing indirect effects of action applications, axioms facilitate the
modelling of planning domains and can lead to a more efficient search for
a plan \cite{ivankovic-haslum-ijcai2015}.
The Planning Domain Definition Language PDDL is the dominant language for
specifying classical planning tasks.  A variant of axioms has already been
introduced in PDDL 1.2 \cite{mcdermott-et-al-tr1998} but has been dropped again
in PDDL 2.1 \cite{fox-long-jair2003} because the semantics were not clear. The
previous example axiom corresponds to the form that was reintroduced in PDDL
2.2 \cite{edelkamp-hoffmann-tr2004} and is still in effect in today's PDDL
standard. This last revision is backed up by work by
\inlinecite{thiebaux-et-al-aij2005} that established that axioms increase the
expressive power of PDDL and cannot be compiled away without a worst-case
super-polynomial increase of the task representation or plan size.

Interestingly, the official definition differs from the variant considered in that paper in
an aspect, namely by the restriction on negative occurrences of predicates in axiom
bodies. We call the axiom formalism from the official PDDL definition
$\semposaxioms$ and the more general one from the compilability analysis
$\genaxioms$. In $\semposaxioms$, only basic predicates may occur negatively in
axiom bodies, whereas $\genaxioms$ also permits negative occurrences of derived
predicates as long as the set of axioms is \emph{stratifiable}. This concept
allows to partition the axioms into several strata that are successively
evaluated by individual fixed-point computations. A derived predicate may occur
negatively in the body of an axiom if its interpretation has already been
finalized by an earlier stratum.

Consider as an example in addition to axiom \eqref{eq:path} an axiom
\begin{equation}
  \pddlaxiom{\pacyclic()}{\forall x \lnot \ppath(x,x)}.
  \label{eq:acyclic}
\end{equation}

The negative occurrence of derived predicate $\ppath$ in this axiom would be
permitted in $\genaxioms$ but not in $\semposaxioms$.

\inlinecite{thiebaux-et-al-aij2005} were aware of this difference of axiom
formalisms but did not consider this restriction on the axiom language in their
analysis.

The literature considers both variants.
Some works adopt the restrictions of the PDDL standard and use $\semposaxioms$
\cite{gazen-knoblock-ecp1997,coles-smith-jair2007, gerevini-et-al-amai2011},
while others rely on the more general concept of stratifiability
and consider $\genaxioms$
\cite{helmert-aij2009,borgwardt-et-al-aaai2022,grundke-et-al-icaps2024}.

The planning literature also considers planning tasks in a propositional
or finite-domain representation (FDR), with \emph{ground} actions and axioms
(i.e.\ axioms do not mention any variables). In this context, it is prevalent
to permit (stratifiable) negative occurrences of derived predicates
\cite{ivankovic-haslum-ijcai2015}, although sometimes only restricted to an FDR
variant of (ground) stratified Datalog
\cite{miura-fukunaga-icaps2017,speck-et-al-icaps2019,speck-gnad-icaps2024}.
Since we focus on lifted axiom formalisms, these formalisms are not in the
scope of our analysis.

In this paper, we make the following contributions:
\begin{itemize}
  \item In Section \ref{sec:equivalence}, we show that both variants  $\genaxioms$ and $\semposaxioms$ can
    express exactly the same queries as least fixed point logic.
  \item Based on this insight, we clarify their relationship to stratified
    Datalog in Section \ref{sec:datalog}.
  \item The theoretical result indicates that negative occurrences of derived
    predicates can be eliminated. We present the corresponding compilation
    in Section \ref{sec:elimination}.
\end{itemize}

Since we build on known results for least fixed point logic, we first provide
the necessary background, formally introducing this logic and the axiom
formalisms $\genaxioms$ and $\semposaxioms$.

\section{Background}
\label{sec:background}

We assume that the reader is familiar with first-order logic (\FO). 
We only consider first-order vocabularies with a finite set of predicates and 
constants and without function symbols (besides the constants).
We also only consider \emph{finite} structures $\struct$ that interpret all
predicates and constant symbols for a finite universe $U$.

We write $\varphi(x_1,\dots,x_n)$ to indicate that $x_1,\dots,x_n$ are the free
variables in formula $\varphi$. A variable assignment maps the free variables to
elements of the universe. We write $\struct\models \varphi(o_1,\dots,o_n)$ to
indicate that $\varphi(x_1,\dots,x_n)$ is true under $\struct$ and the variable
assignment that maps $x_i$ to $o_i\in U$, using the usual semantics of $\FO$.
We also write $\vec x$ and $\vec o$ for vectors of variables and objects,
leaving their length implicit.

Fixed point logic can define relations that are not already interpreted by the
given structure.  We write $\varphi(P_1,\dots,P_m, x_1,\dots,x_n)$ to indicate
that predicates $P_1,\dots,P_m$ in formula $\varphi$ refer to such relations
(and $x_1,\dots,x_n$ are free variables).

An occurrence of a predicate in a formula is \emph{positive} if it is
under the scope of an even number of negations. Otherwise, it is
\emph{negative}. For example, in $\exists x \lnot P(x) \land\lnot\forall
y\exists z \lnot( P(y) \lor \lnot P(z))$ the first occurrence of $P$ (i.e.\
$P(x)$) is negative, the second one ($P(y)$) positive, and the last one
($P(z)$) again negative. In the planning literature
\cite{thiebaux-et-al-aij2005,edelkamp-hoffmann-tr2004}, the same concept of
negative occurrences is also described as negated appearances in the negation
normal form of $\varphi$.

\subsection{Expressiveness}

We will compare the expressiveness of different formalisms
based on which \emph{queries} they can express. A query associates a structure
$\mathcal S$ (with universe $U$) with a subset of $U\times\dots\times U$.

For logic formalisms such as least fixed point logic, a query is given by a
formula $\varphi(x_1,\dots,x_n)$ with free variables $x_1,\dots,x_n$ and
defines the mapping
$\varphi(\struct) =
\{\langle o_1,\dots,o_n\rangle\in U^n\mid
\struct\models\varphi(o_1,\dots,o_n)\}.\footnote{We follow
\inlinecite{libkin-2004} in overloading notation.}$

We will later also introduce a form of queries for the axiom formalisms under
study.

For two formalisms $X$ and $Y$, we write $X\leq Y$ to express that every
$X$-query $Q_X$ has an equivalent $Y$-query $Q_Y$, i.e.\ both associate every
given finite structure $\struct$ with the same set.
In that case we say that $Y$ is \emph{at least as expressive} as $X$. It is
easy to see that $\leq$ is transitive. We write $X = Y$ and call them
\emph{equally expressive} if $X\leq Y$ and $Y\leq X$, which means that both
formalisms can define the same queries. A formalism $Y$ is \emph{strictly more
expressive} than a formalism $X$, written $X < Y$, if $X\leq Y$ and $X\neq Y$.

\subsection{Fixed Points}
The formal definition of least fixed point logic and the semantics of PDDL
axioms rely on the general concept of fixed points.
Fixed points are a property of operators of the form $F:\mathcal
P(S)\rightarrow \mathcal P(S)$, where $S$ is a set and $\mathcal P(S)$
denotes its power set. An operator $F:\mathcal P(S)\rightarrow \mathcal P(S)$ is
called
\begin{itemize}
  \item \emph{monotone} if $X\subseteq Y$ implies $F(X)\subseteq F(Y)$, and
  \item \emph{inductive} if the sequence $X_0 = \emptyset$, $X_{i+1} = F(X_i)$
is increasing, i.e.\ $X_i\subseteq X_{i+1}$ for all $i$.
\end{itemize}

Every monotone operator is inductive and for inductive operators we define
$X_\infty = \bigcup_{i=0}^\infty X_i$. We only consider finite sets $S$, so there
always is an $n\in\mathbb N_0$ with $X_n = X_\infty$.

We call a set $X\subseteq S$ a \emph{fixed point} of operator $F$ if $X=F(X)$.
It is the \emph{least} fixed point $\lfp(F)$ if every other fixed point $Y$ of
$F$ is a superset of $X$. We will later use the fact that every monotone
operator $F$ has a least fixed point $\lfp(F)=\bigcap\{X\mid X = F(X)\}$ and
that $\lfp(F) = X_\infty$ (e.g.\ \citeauthor{libkin-2004}
\citeyear{libkin-2004}, Theorem 10.2).

\subsection{Least Fixed Point Logic}

In least fixed point logic (\LFP) we can introduce a new relation as the least
fixed point of an operator that is defined in terms of a formula.

Let $P$ be a new predicate symbol of arity $k$, i.e.\ $P$ is not interpreted by
$\struct$, and let $\varphi(P,x_1,\dots,x_k)$ be a formula that is positive in $P$.

It induces an operator $F_\varphi:\mathcal P(U^k)\rightarrow\mathcal
P(U^k)$ defined as $$F_\varphi(X) = \{\langle o_1,\dots,o_k\rangle\mid
\struct \models\varphi(P/X,o_1,\dots,o_k)\},$$ where $P/X$ means that $P$ is
interpreted as $X$ in $\varphi$.
If $\varphi(P,x_1,\dots,x_k)$ is positive in $P$ then $F_\varphi$ is monotone
\cite[Lemma 10.7]{libkin-2004}.

For illustration, we demonstrate the computation of the least fixed point of $F_\varphi$ for
the example formula
\begin{equation*}
\varphi(\ppath,x,y) := E(x,y)\lor \exists
z(E(x,z)\land \ppath(z,y))
\end{equation*}
and a structure $\struct$ with universe $U = \{a,b,c,d\}$ which interprets $E$
as $\{\langle a,b\rangle, \langle b,c\rangle, \langle c,d\rangle\}$.
Starting from $X_0 = \emptyset$, we collect all tuples $\langle o_1,o_2\rangle \in U\times
U$ for which $\struct \models \varphi(\ppath/\emptyset,o_1,o_2)$ holds. This is
the case only for the tuples for which $E$ is true.  The existential subformula
of $\varphi$ is false for all $z\in U$ because $\ppath$ is interpreted as the
empty set $X_0$ and thus false for all tuples.  So, we get $X_1:=F_\varphi(X_0)
= \{\langle a,b\rangle, \langle b,c\rangle, \langle c,d\rangle\}$.

Applying the operator $F_\varphi$ to $X_1$, we get $X_2 := F_\varphi(X_1) =
\{\langle a,b\rangle, \langle b,c\rangle, \langle c,d\rangle,\langle
a,c\rangle,\langle b, d\rangle\}$. For instance, $\langle a,c\rangle$ is
included in $X_2$ because $\struct \models \varphi(\ppath/X_1,a,c)$ holds,
i.e.\ because $U$ contains element $b$, for which $E(a,b)$ is true (interpreted
by $\struct$) and $\ppath(b,c)$ is true ($\ppath$ being interpreted as $X_1$).
If we continue the sequence, we get $X_3 = X_2\cup\{\langle a,d\rangle\}$ and
$X_4=X_3$. The sequence reached a fixed point and we know that
$X_\infty=\lfp(F_\varphi)= X_3$.

We are now prepared to define the logic $\LFP$:

\begin{definition}[\citeauthor{libkin-2004} \citeyear{libkin-2004}, Definition 10.6]
The logic $\LFP$ extends $\FO$ with the following formation rule:
\begin{itemize}
\item If $\varphi(P,\vec{x})$ is a formula positive in $P$, where $P$ is
$k$-ary, and $\vec t$ is a tuple of terms, where $|\vec{x}| = |\vec{t}| = k$,
then \[[\lfp_{P,\vec{x}}\varphi(P,\vec{x})](\vec t)\]
is a formula, whose free variables are those of $\vec{t}$.
\end{itemize}
Semantically, $\struct\models[\lfp_{P,\vec{x}}\varphi(P,\vec{x})](\vec o)$ iff
  $\vec{o}\in\lfp(F_\varphi)$.
\label{def:LFP}
\end{definition}

Continuing the graph example, we can see that the $\LFP$ formula
$[\lfp_{\ppath, x,y}\varphi(\ppath,x,y)](a,d)$ holds in $\struct$ because
$\langle a,d\rangle\in X_3 = X_\infty$. As $X_\infty$ is irreflexive, $\struct$
also satisfies formula $\lnot\exists z [\lfp_{\ppath,
x,y}\varphi(\ppath,x,y)](z,z)$, expressing that a graph with edges as in the
interpretation of $E$ by $\struct$ is acyclic.

\subsubsection{Simultaneous Fixed Points}

In the previous definitions, the fixed point only iterates a single predicate.
The formalism can be extended to simultaneous fixed points as follows:

Let $P_1,\dots,P_n$ be new predicate symbols, where the arity of $P_i$ is $k_i$
and let $\vec{x}_i$ be a $k_i$-tuple of variables. Consider a set
$\Phi=\{\varphi_1(P_1,\dots,P_n,\vec{x}_1),\dots,\varphi_n(P_1,\dots,P_n,\vec{x}_n)\}$
of formulas that are positive in all predicates $P_i$.

Each formula $\varphi_i$ defines an operator $F_i:\mathcal P(U^{k_1})
\times\dots\times \mathcal P(U^{k_n})\rightarrow \mathcal P(U^{k_i})$ as
$$F_i(X_1,\dots,X_n) = \{\vec{o}\mid \struct
\models\varphi_i(P_1/X_1,\dots,P_n/X_n,\vec{o})\}.$$

These operators induce a sequence of vectors from $\mathcal
P(U^{k_1})\times\dots\times \mathcal P(U^{k_n})$ by $\vec X_0
= \langle\emptyset,\dots,\emptyset\rangle$ and $\vec X_{i+1} = \langle F_1(\vec
X_i),\dots,F_n(\vec X_i)\rangle$. This sequence has a fixed point, which we
denote by $\vec X_\infty$.

The logic $\LFPsim$ extends $\FO$ with the formation rule
$[\lfp_{P_i,\Phi}](\vec t)$, where $\vec t$ is a tuple of length $k_i$. The
semantics is that $\struct\models[\lfp_{P_i,\Phi}](\vec o)$ iff $\vec o$ is an
element of the $i$-th component of $\vec X_\infty$.

Consider for example $P_1$ and $P_2$ with arity $1$ and $2$, respectively, and $\Phi
= \{Q(x)\lor\exists y P_2(y,x), P_1(x)\land R(x,y)\}$. For $\struct$ with
universe $\{a,b,c,d\}$ such that exactly $R(a,b)$, $R(b,c)$, $R(c,d)$ and $Q(b)$ are
true, we get the following sequence for the $\vec X_i$:
$\vec X_1 = \langle \{\langle b\rangle\},\emptyset\rangle$, 
$\vec X_2 =$ $\langle \{\langle b\rangle\},\{\langle b,c\rangle\}\rangle$,\\
$\vec X_3 = \langle \{\langle b\rangle, \langle c\rangle\},\{\langle b,c\rangle\}\rangle$, 
$\vec X_4 = \langle \{\langle b\rangle, \langle c\rangle\},\{\langle
b,c\rangle,$ $\langle c,d\rangle\}\rangle$, $\vec X_5 = \langle \{\langle
b\rangle, \langle c\rangle, \langle d\rangle\},\{\langle b,c\rangle,$ $\langle
c,d\rangle\}\rangle
= \vec X_6$. So for instance, $\struct \models [\lfp_{P_1,\Phi}]\langle b\rangle$ and
$\struct \models \lnot[\lfp_{P_2,\Phi}](a, b)$.

Simultaneous iteration of several predicates does not increase the expressive
power compared to the iteration of a single predicate:

\begin{theorem}[\inlinecite{libkin-2004}, Cor.~10.8]
  $\LFPsim=\LFP$.
  \label{thm:lfpsim}
\end{theorem}

\subsection{Axiom Programs}
A planning domain defines a finite relational \FO-vocabulary (i.e.\ one
containing only predicates and constants), a set of actions and an axiom
program. We skip the specifics of the actions as they are not relevant for this
work. Instead we focus on the predicates, states and axioms. 
The predicates are partitioned into the \emph{basic} and \emph{derived}
predicates. The derived predicates are those that occur in the heads of axioms.
In the context of our work, we treat states as
structures that interpret the predicates over a fixed universe
$U$, consisting of the objects of the planning task. A state can alternatively
be seen as the set of true ground atoms or a truth assignment to the ground
atoms.

A \emph{basic} state (or \emph{basic} structure) only considers the basic
predicates, whose interpretation is directly defined by the action applications.
It gets extended to a full state (considering all predicates) by means of the
axioms.

\begin{definition}[Axiom]
  An \emph{axiom} has the form $$\pddlaxiom{P(\vec x)}{\varphi(\vec x)},$$ where
  $P(\vec x)$ is a $\FO$ atom and $\varphi(\vec x)$ is a $\FO$
  formula such that $P(\vec x)$ and $\varphi(\vec x)$ have the same
  free variables $\vec x$. 

  We call $P(\vec x)$ the \emph{head} and $\varphi(\vec x)$ the \emph{body} of
  the axiom. The axiom \emph{affects} the head predicate $P$.
  A set $\Pi$ of axioms \emph{affects} $P$ if some axiom in $\Pi$ affects $P$.
\end{definition}

The most general type of axiom programs commonly considered in planning are
\emph{stratifiable} axiom programs that enable a well-defined semantics.
For ease of presentation, we directly require a specific stratification. This
does not limit the scope of our work because all stratifiable axiom programs
can be represented in this form and all stratifications of a stratifiable
program are semantically equivalent \cite[Thm.~11]{apt-et-al-1988}.  Our
definition is in this respect analogous to the definition of stratified Datalog
by \inlinecite{ebbinghaus-flum-1995}.
Intuitively, stratification prohibits recursion through negation.

\begin{definition}[Stratified Axiom Program, $\genaxioms$]
  A \emph{stratified axiom program} is a finite sequence
  $\langle\Pi_1,\dots,\Pi_n\rangle$, where
  \begin{itemize}
    \item each $\Pi_i$ is a finite set of axioms, called a \emph{stratum},
    \item all axioms affecting the same predicate $P$ are in the same stratum,
      and
    \item for all strata $\Pi_i$ and $\pddlaxiom{P(\vec x)}{\varphi(\vec x)}$ in $\Pi_i$ it holds that 
    \begin{itemize}
      \item if a predicate $Q$ appears positively in $\varphi(\vec x)$ then the axioms
        affecting $Q$ are in $\bigcup_{k=1}^i \Pi_k$, and
      \item if a predicate $Q$ appears negatively in $\varphi(\vec x)$ then the axioms
        affecting $Q$ are in $\bigcup_{k=1}^{i-1} \Pi_k$.
    \end{itemize}
  \end{itemize}
  \noindent We refer to the class of all stratified axiom programs as $\genaxioms$.
\end{definition}

The PDDL standard \cite{edelkamp-hoffmann-tr2004} only permits negative
occurrences of \emph{basic} predicates in axiom bodies.
Under this restriction all axioms can be combined into a single stratum
\cite{thiebaux-et-al-aij2005} and every stratified axiom program with a single
stratum respects this restriction. We use this property to define the second
axiom formalism that we want to consider.

We call such axiom programs \emph{semipositive}, following the terminology for
the analogous restriction in Datalog with negation \cite{abiteboul-et-al-1995}.

\begin{definition}[Semipositive Axiom Program, $\semposaxioms$]

  A \emph{semipositive} axiom program is a stratified axiom program
  $\langle\Pi\rangle$ that consists of a single stratum.
  We refer to the class of all semipositive axiom programs as $\semposaxioms$.
\end{definition}

For the semantics of axiom programs, we first consider a single axiom stratum
$\Pi$. Let $P_1,\dots,P_m$ with arities $k_1,\dots,k_m$ be the predicates
affected by the axioms in $\Pi$ and let $\struct$ be a structure that interprets
all predicates in $\Pi$ except for $P_1,\dots,P_m$ on some finite universe $U$.
In the overall context, for the first stratum, $\struct$ will be a basic state.
For later strata, it will in addition interpret all predicates affected by
earlier strata in the axiom program.

Stratum $\Pi$ extends $\struct$ to a structure $\struct[\Pi]$ that also
interprets the predicates $P_1,\dots,P_m$ affected by $\Pi$, while preserving
the interpretation of all other symbols from $\struct$.

Conceptually, we consider all ground instantiations of the axioms in the
stratum and make the head true if the body is true under $\struct$ plus the
already derived head atoms until we reach a fixed point. The interpretation of
$P_i$ is then the set of all derived ground atoms with predicate $P_i$.

Formally, the interpretation for $P_1,\dots,P_m$ can be defined by the least
simultaneous fixed point of the operator $F:\mathcal P(U^{k_1})\times\dots\times
\mathcal P(U^{k_m})\rightarrow \mathcal P(U^{k_1})\times\dots\times \mathcal
P(U^{k_m})$,
where
$F(X_1,\dots,X_m) =
  \langle F_1(X_1,\dots,X_m), \dots,F_m(X_1,\dots,X_m)\rangle$
with
\begin{multline*}
$$F_i(X_1,\dots,X_m) =\\ \bigcup_{\mathclap{\pddlaxiom{P_i(\vec
x)}{\varphi(P_1,\dots,P_m,\vec x)}\in \Pi}}\{\vec o \mid
\struct\models
\varphi(P_1/X_1,\dots,P_m/X_m,\vec o)\}.
\end{multline*}

If stratum $\Pi$ contains for derived predicate $P_i$ only one axiom $\pddlaxiom{P_i(\vec x)}{\varphi(P_1,\dots,P_m,\vec
x)}$ then $F_i$ simplifies to $F_i(X_1,\dots,X_m) =
\{\vec o \mid \struct\models\varphi(P_1/X_1,\dots,P_m/X_m,\vec o)\}$.

Since $F$ is monotone, the sequence of vectors from $\mathcal
P(U^{k_1})\times\dots\times P(U^{k_m})$ defined by $\vec X_0
= \langle\emptyset,\dots,\emptyset\rangle$ and $\vec X_{i+1} = F(\vec X_i)$
reaches a fixed point $\vec X_\infty$ that corresponds to the least fixed point
of $F$.

The $i$-th component of this least fixed point defines the interpretation of
predicate $P_i$ in $\struct[\Pi]$.

An entire stratified axiom program $\prog = (\Pi_1,\dots,\Pi_n)$ extends
a basic state $\struct$ stratum by stratum to $$\struct[\prog] = (\dots
(\struct[\Pi_1])[\Pi_2]\dots)[\Pi_n].$$

We can now turn to the question whether $\genaxioms$ is strictly more expressive
than $\semposaxioms$ and how both compare to $\LFP$.

\section{Equivalence of $\genaxioms$ and $\semposaxioms$ to $\LFP$}
\label{sec:equivalence}
To compare the expressiveness of the class of stratified axiom programs to least
fixed point logic, we need to define the notion of a query for this formalism.
For the related Datalog formalism, \inlinecite{ebbinghaus-flum-1995} define
a concept of Datalog formulas. We use the same concept for axioms queries.

\begin{definition}[Axiom query]
  An \emph{axiom query} is a pair
  $\axquery=\langle\prog, P\rangle$, where $\prog$ is a stratified axiom
  program and $P$ is a predicate with arity $k$ affected by some stratum of
  $\prog$.

  For basic structure $\struct$ with universe $U$, 
  $$\axquery(\struct) = \{\langle o_1,\dots,o_k\rangle\in U^k\mid \struct[\prog]
  \models P(o_1,\dots,o_k)\}.$$
\end{definition}

The query determines which ground atoms for predicate $P$ can be derived from
a given basic state.

Since semipositive axiom programs are a special case of
stratified axiom programs, our first theorem is trivial.

\begin{theorem}
  $\semposaxioms\leq\genaxioms$ 
  \label{thm:semposgen}
\end{theorem}

Overall, we will establish that indeed $\semposaxioms=\genaxioms=\LFP$.

For this purpose, we consider a syntactically very limited fragment of
$\LFPsim$ called $\LFP_0$ that extends $\FO$ as follows: Let $\Phi$ be a finite
set of $\FO$ formulas $\varphi(P_1,\dots,P_n,\vec{x})$ that are positive in all
$P_i$. Then $[\lfp_{P_i,\Phi}](\vec{x})$ is an $\LFP_0$ formula. Note that this
is not a formation rule as in Definition \ref{def:LFP}, so these formulas may
not be further combined with Boolean connectives, quantification or fixed point
operators.

Although $\LFP_0$ is syntactically very restrictive, it can be shown that it is
as expressive as full $\LFP$:

\begin{theorem}[\inlinecite{libkin-2004}, Cor.~10.13]
  $\LFP_0=\LFP$.
  \label{thm:lfp0}
\end{theorem}

We will separately establish that $\LFP_0\leq \semposaxioms$ (Theorem
\ref{thm:lfp0sempos}) and that $\genaxioms\leq\LFPsim$ (Theorem
\ref{thm:genlfpsim}).  With Theorem \ref{thm:semposgen} and $\LFPsim = \LFP
= \LFP_0$ from Theorems \ref{thm:lfpsim} and \ref{thm:lfp0}, this implies that
$\LFP \leq\semposaxioms \leq \genaxioms\leq \LFP$, establishing the equality of
the formalisms.

\begin{theorem}
$\LFP_0\leq \semposaxioms$.
  \label{thm:lfp0sempos}
\end{theorem}

\begin{proof}
  We show how we can translate every $\LFP_0$ query $\varphi(\vec x)$ into an
  equivalent $\semposaxioms$ query $\axquery = (\prog,Q)$.
  \smallskip

  Let $\varphi(\vec x)$ be an arbitrary $\LFP_0$ formula.
  Formula $\varphi(\vec x)$ is either a $\FO$ formula or it has the form
  $[\lfp_{P_\ell,\Phi}](\vec{x})$, where $\Phi$ is a finite set of $\FO$
  formulas.
  \smallskip

  If $\varphi(\vec x)$ is a $\FO$ formula, we introduce a new predicate $Q$ with
  arity $|\vec{x}|$ and use an axiom program
  $$\prog=\langle\{\pddlaxiom{Q(\vec{x})}{\varphi(\vec x)}\}\rangle$$ with a single axiom.
  Since $Q$ is the only derived predicate and $\varphi(\vec x)$ does not mention
  it, this program is trivially stratified. Since it consists of a single
  stratum, it is semipositive. It is easy to see that the query $\axquery
  = \langle\prog, Q\rangle$ is equivalent to the query $\varphi(\vec x)$.

  If $\varphi(\vec x)$ has the form $[\lfp_{P_\ell,\Phi}](\vec{x})$ with
  $\Phi=\{\varphi_1(P_1,\dots,P_n,\vec{x}_1),\dots,\varphi_n(P_1,\dots,P_n,\vec{x}_n)\}$,
  we construct a program with a single stratum
  $\Pi$ as follows:

  For each formula $\varphi_i(P_1,\dots,P_n,\vec{x}_i)\in \Phi$, stratum $\Pi$
  contains an axiom
  $\pddlaxiom{P_i(\vec{x}_i)}{\varphi_i(P_1,\dots,P_n,\vec{x}_i)}$. Note that
  for $\varphi$ being an $\LFP_0$ formula, each $\varphi_i$ must be positive in
  all predicates $P_1,\dots,P_n$, so the axiom program $\prog:=
  \langle\Pi\rangle$ is stratified and semipositive. The $\semposaxioms$ query
  $\langle\prog,P_\ell\rangle$ is equivalent to $\varphi(\vec x)$.
\end{proof}

\begin{theorem}
  $\genaxioms\leq\LFPsim$.
  \label{thm:genlfpsim}
\end{theorem}

\begin{proof}
  Let $\axquery=\langle\prog,Q\rangle$ be an $\genaxioms$ query with
  $\prog=\langle\Pi_1,\dots,\Pi_n\rangle$. We show how we can translate
  $\axquery$ into an equivalent $\LFPsim$ formula $\varphi$.

  We proof the statement by an induction over the strata.

  We first only consider the first stratum $\Pi_1$. Let $P_1,\dots,P_m$ be the
  derived predicates affected by this stratum. In a first step, we combine all
  axioms affecting the same predicate into a single axiom.
  For $i\in\{1,\dots,m\}$, let $\Pi^{P_i}_1\subseteq \Pi_1$ consist of the
  axioms that affect $P_i$.  We can assume w.l.o.g.\
  that all these axioms have the form $\pddlaxiom{P_i(\vec
  x)}{\varphi(P_1,\dots,P_n,\vec x)}$, using the same vector $\vec x$ of free
  variables. Otherwise we can rename the variables accordingly, if necessary
  replacing bound variables with fresh variables.

  For each derived predicate $P_i$, let
  \begin{equation}
    \psi_{P_i}(P_1,\dots,P_m,\vec {x}) := \bigvee_{\qquad\mathclap{\pddlaxiom{P_i(\vec x)}{\varphi(P_1,\dots,P_m,\vec
    x)}\in\Pi^{P_i}_1}\qquad}\varphi(P_1,\dots,P_m,\vec x)
    \label{eq:psiPi}
  \end{equation}
  and define formula set $\Psi := \{\psi_{P_i}(P_1,\dots,P_m,\vec {x})\mid
  i\in\{1,\dots,m\}\}$.

  Since $\Pi_1$ is an axiom stratum, all bodies in its axioms are positive in
  $P_1,\dots,P_m$ and consequently also all formulas
  $\psi_{P_i}(P_1,\dots,P_m,\vec {x})$ are positive in all these predicates.

  Thus $\chi_{P_i}(\vec t):=[\lfp_{P_i,\Psi}](\vec t)$ is an $\LFPsim$ formula.
  It has the property that for all structures $\struct$ it holds that
  $\langle\langle\Pi_1\rangle,P_i\rangle(\struct) = \chi_{P_i}(\struct)$, which can 
  be seen from the semantics of axiom programs and least fixed point logic.
  Both induce the same operator for the simultaneous fixed point computation.
  The crucial insight is that the union in
  \begin{multline*}
  F_i(X_1,\dots,X_m) =\\ \bigcup_{\mathclap{\pddlaxiom{P_i(\vec
  x)}{\varphi(P_1,\dots,P_m,\vec x)}\in \Pi}}\{\vec o \mid
  \struct\models
  \varphi(P_1/X_1,\dots,P_m/X_m,\vec o)\}.
  \end{multline*}
  from the semantics of an axiom stratum is matched by the disjunction in
  equation \eqref{eq:psiPi} for $\psi_{P_i}$. Thus, $F_i$ is equivalent to
  $F_i(X_1,\dots,X_m)=\{\vec{o}\mid \struct
  \models\psi_{P_i}(P_1/X_1,\dots,P_m/X_m,\vec{o})\}$ which is the
  corresponding function in the definition of the simultaneous fixed point in
  $\LFPsim$.
  Since later strata do not change the interpretation of $P_i$,
  it holds also for all other $j$ with $1<j\leq n$ that
  $\langle\langle\Pi_1,\dots,\Pi_j\rangle,P_i\rangle(\struct)=
  \chi_{P_i}(\struct)$. 
  \medskip
 
  For the induction hypothesis, suppose that for every derived predicate $P$
  affected by an axiom stratum $\Pi_\ell$ with $\ell<l$, there is an $\LFPsim$
  formula $\chi_{P}(\vec t)$ such that for all $\ell\leq j\leq n$ it holds for
  all structures $\struct$ that
  $\langle\langle\Pi_1,\dots,\Pi_j\rangle,P\rangle(\struct)=\chi_{P}(\struct)$.
  \medskip

  We show that then there is also such a formula $\chi_P$ for every 
  predicate $P$ affected by stratum $\Pi_l$. For this purpose, we construct for each
  such $P$ a formula $\psi_P$ as in equation \ref{eq:psiPi} with
  the difference that for every occurrence of an atom $P'(\vec t')$ where $P'$
  is affected by an earlier stratum, we use instead formula
  $\chi_{P'}(\vec t')$ from the induction hypothesis. With $\Psi_l:=\{\psi_P\mid
  P\text{ is affected by }\Pi_l\}$, we can define $\LFPsim$-formula $\chi_{P}(\vec t)
  := [\lfp_{P,\Psi_l}](\vec t)$ as the desired formula with
  $\langle\langle\Pi_1,\dots,\Pi_j\rangle,P\rangle(\struct)= \chi_{P}(\struct)$ for all $l\leq j\leq
  n$ for all structures $\struct$.

  We have shown that for every derived predicate $P$ there is an
  equivalent $\LFPsim$ formula $\chi_P$ with
  $\langle\prog,P\rangle(\struct)=\chi_P(\struct)$. So this is also true for
  predicate $Q$ and we have overall shown that for every $\genaxioms$ query
  there is an equivalent $\LFPsim$ query.
\end{proof}

Consider as an example an $\genaxioms$ program $\prog$ with axioms
\begin{align}
  \pddlaxiom{P(x,y,z)}{&E(x,y)\land(x\neq y)\land
  (y\neq z)}\label{example:P1}\\
  \pddlaxiom{P(x,y,z)}{&\exists u(E(x,u)\land
  P(u,y,z)\land(x\neq z))}\label{example:P2}\\
  \pddlaxiom{Q(x,y)}{&E(x,y)\lor{}\label{example:Q}\\&\exists u(R(x,u)\land E(u,y)\land
  P(x,u,y))}\nonumber\\
  \pddlaxiom{R(x,y)}{&\exists u(Q(x,u)\land Q(u,y)\land P(x,u,y))}\label{example:R}\\
  \pddlaxiom{S(x)}{&\lnot Q(x,x)}\label{example:S},
\end{align}
with strata
$\langle\Pi_1=\{\eqref{example:P1},\eqref{example:P2}\},\Pi_2=\{\eqref{example:Q},\eqref{example:R}\},\Pi_3=\{\eqref{example:S}\}\rangle$.
For stratum $\Pi_1$, we use formula set $\Psi_1 = \{\varphi_P(P,x,y,z)\}$ with
$\varphi_P(P,x,y,z)=(E(x,y)\land(x\neq y)\land (y\neq z))\lor\exists u(E(x,u)\land
P(u,y,z)\land(x\neq z))$.
For stratum $\Pi_2$, we use set $\Psi_2
= \{\varphi_Q(Q,R,x,y),\varphi_R(Q,R,x,y)\}$ with
$\varphi_Q(Q,R,x,y)=E(x,y)\lor\exists u(R(x,u)\land E(u,y)\land
  [\lfp_{P,\Psi_1}](x,u,y))$ and 
$\varphi_R(Q,R,x,y) = \exists u(Q(x,u)\land Q(u,y)\land [\lfp_{P,\Psi_1}](x,u,y))$. For
the last stratum, we get $\Psi_3 = \{\varphi_S(x)\}$ with $\varphi_S(x)
= \lnot[\lfp_{Q,\Psi_2}](x,x)$. Now for all $\struct$ (interpreting $E$), 
$([\lfp_{Q,\Psi_3}](x))(\struct) = \langle\prog, Q\rangle(\struct)$.

With the discussion at the beginning of this section we get the
main result of this paper as a corollary of Theorems \ref{thm:semposgen},
\ref{thm:lfp0sempos} and \ref{thm:genlfpsim}:

\begin{corollary}
  $\genaxioms = \semposaxioms = \LFP$.
  \label{corr:apisap0}
\end{corollary}

In the next section, we will use this result to relate $\genaxioms$ and
$\semposaxioms$ to another frequently studied axiom formalism in planning,
addressing a common misconception.

\section{Relationship to Stratified Datalog}
\label{sec:datalog}

$\LFP$ is a well-studied formalism. For instance, it is known that
\emph{bounded} fixed-point logic (also known as stratified fixed-point logic)
$\BFP$ is strictly weaker than $\LFP$ \cite[Th.\ 7.7.2]{ebbinghaus-flum-1995}.
This logic $\BFP$ in turn is equally expressive as \emph{stratified Datalog}
\cite[Th.\ 8.1.1]{ebbinghaus-flum-1995}, which is a formalism very similar to
$\genaxioms$, but restricting the bodies to (implicitly) existentially
quantified \emph{conjunctions} of literals.

This result on stratified Datalog is relevant in the context of axioms in
planning. The Fast Downward planning system \cite{helmert-jair2006} translates
the input task in a preprocessing phase to a normal form where axioms are
stratified Datalog rules \cite{helmert-aij2009}. 
The difference in expressiveness reveals a fundamental limitation of this
translation because according to our results, it is not always possible.
Indeed, a closer look at Fast Downward's transformation reveals that it can
lead to non-stratifiable programs.

As an example, consider axiom $A:= \pddlaxiom{S(x)}{\exists y(M(x,y)\land
\forall z (M(y,z) \rightarrow S(z)))}$. This example is taken from the
proof by \inlinecite{kolaitis-iandc1991} where he establishes that fixed point logic
has a higher expressive power than stratified Datalog by means of a game-tree
separation. The axiom expresses that player 1 has a winning strategy from
position $x$ if she can move to a position $y$ such that for all possible moves from
$y$ to $z$ by player 2, player 1 has a winning strategy from $z$.
The transformation by Fast Downward would introduce a new predicate $R(y)$ with axiom
$B:=\pddlaxiom{R(y)}{\exists z(M(y,z) \land \lnot S(z))}$ and replace the axiom $A$ with
$A':=\pddlaxiom{S(x)}{\exists y(M(x,y)\land \lnot R(y))}$. Now $S$ occurs
negatively in the body of axiom $B$, so axiom $A'$ must be in a strictly
earlier stratum than $B$. But since $R$ has a negative occurrence in the body
of axiom $A'$, axiom $B$ must be on a strictly earlier stratum than $A'$.
Overall, we see that a stratification is no longer possible after this step of
the transformation. In this example, the problem occurs because the head
predicate $S$ occurs again under the scope of an universal quantifier in the
body.

While a general transformation from $\genaxioms$ to stratified Datalog is not
possible according to our theoretical results, the transformation by Fast
Downward must only be correct for a specific given task, which has a fixed
finite universe. Thus it would be possible to expand universal quantifiers in
axiom bodies for the entire universe (for the prize of a blow-up in the
representation size). It could also be possible to apply a transformation that
shifts part of the axiom evaluations into forced action applications, similar
to the compilations by \inlinecite{thiebaux-et-al-aij2005}.

Also the work by \inlinecite{thiebaux-et-al-aij2005} needs to be revisited in
the light of our theoretical findings. They write:

``\emph{Furthermore, we will often restrict the form of the right-hand
side of axioms to a particularly simple form, namely, conjunctions of atoms, where
all variables not appearing on the left-hand side of the axiom are implicitly
existentially quantified. Such axioms are syntactically identical to DATALOG
programs and for this reason we will call such axioms to be in DATALOG form. If
the axioms contain negation, we say that they are in DATALOG$^{\lnot}$ form. It
is now a well-known fact from database theory that first-order queries can be
rewritten into DATALOG$^{\lnot}$ programs, which are linear in the size of the
original formula \cite{abiteboul-et-al-1995}. For this reason, we can
concentrate in the following on stratified axioms in DATALOG$^{\lnot}$ form as
the most expressive axiom language.}''

According to our theoretical result, this formalism is strictly less expressive
than general PDDL axiom programs, so something must be wrong with the
rewriting argument.

Indeed, FO queries can be rewritten to non-recursive Datalog with negation.
This is a special case of stratified Datalog, where the definition of no
predicate depends (also indirectly) on itself.

The argument that arbitrary first-order formulas $\varphi$ in axiom bodies can
be replaced by a set of axioms that evaluate $\varphi$ is tempting, but there
are two issues: First, the rewriting can require several strata, but we need to
be able to put all these axioms into the same stratum as the rewritten axiom body.
Second, first-order queries have no notion of recursion and the rewriting
evaluates the formula relative to a structure that already interprets
\emph{all} predicates. In this scenario it is permitted to introduce negated
occurrences of predicates, which is not the case in our setting if the
predicate is derived on the same stratum.
Considering again $\pddlaxiom{S(x)}{\exists y(M(x,y)\land \forall z (\lnot M(y,z)
\lor S(z)))}$: the two axioms $\pddlaxiom{Q(x)}{\exists y(M(x,y)\land \lnot
R(y))}$ and $\pddlaxiom{R(y)}{\exists z(M(y,z)\land\lnot S(z))}$ correspond to
such a rewriting of the body and we would combine them with
$\pddlaxiom{S(x)}{Q(x)}$ in one stratum. We see directly that the negated
occurrence of $S$ in the axiom affecting $P$ breaks the stratification.

Note that this insight does not threaten the non-compilability results in the
paper by \citeauthor{thiebaux-et-al-aij2005},
so the central finding that axioms add to the expressive power of PDDL stays
valid.

\section{Eliminating Negative Occurrences}
\label{sec:elimination}

Another implication of our theoretical result is that every $\genaxioms$
program has an equivalent $\semposaxioms$ program, meaning negative occurrences
of derived predicates can be eliminated from axiom bodies. This section
presents such a compilation.

Our theoretical result builds on the fact that $\LFP_0$ and $\LFP$ are
equivalent, which was established by means of an actual compilation from $\LFP$
to $\LFP_0$. We extract its core idea and transfer it to the axiom formalisms
in planning.

The original result for fixed point logic goes back
to Moschovakis \cite{moschovakis-1974} for infinite structures and was adapted
to finite structures by Immerman \cite{immerman-infcontrol1986} and Gurevich
\cite{gurevich-shelah-apal1986}.  We follow the structure by Leivant
\cite{leivant-iandc1990} as presented by Libkin \cite[Cor.\
10.13]{libkin-2004}. This requires as a new contribution to directly handle the
\emph{simultaneous} fixed point within each stratum.

We will transform a given $\genaxioms$ program $\prog = \langle
\Pi_1,\dots,\Pi_n\rangle$ to a program $\prog' = \langle
\Pi'_1,\dots,\Pi'_n\rangle$ that does not contain negative occurrences of
derived predicates. Then all strata $\Pi'_1,\dots,\Pi'_n$ can be combined into
a single stratum, resulting in an equivalent $\semposaxioms$ program.

For every derived predicate $P$ of $\prog$ and every basic structure $\struct$
the structures $\struct[\prog]$ and $\struct[\prog']$ will interpret $P$
equivalently, but program $\prog'$ will use additional derived predicates.
The idea is to introduce for each derived predicate $P$ of $\prog$ a new
derived predicate $\bar P$ of the same arity for its complement, i.e.\
$\struct[\prog']\models \bar{P}(\vec t)$ iff $\struct[\prog'] \not\models P(\vec
t)$ (or equivalently $\struct[\prog]\not\models P(\vec{t})$).
We use these complement predicates to replace negative occurrences $P(\vec x)$
of $P$ with the equivalent $\lnot\bar{P}(\vec x)$. Since $P(\vec x)$ is under the
scope of an odd number of negations, $\bar{P}(\vec x)$ is then under the scope
of an even number, so this is a positive occurrence of $\bar{P}$.

To achieve the intended interpretation of the complement predicates, we will
need a number of auxiliary derived predicates to ``analyze'' what is \emph{not}
derived on each stratum.  

The additional predicates are related to the fixed-point computation for a
single stratum and we introduce them separately for each stratum $\Pi_\ell$ of
$\prog$.

Let $P_1,\dots,P_m$ be the predicates affected by $\Pi_\ell$.
Consider an arbitrary basic structure $\struct$ and the extension of
$\struct[\langle \Pi_1,\dots,\Pi_{\ell-1} \rangle]$ to $\struct[\langle
\Pi_1,\dots,\Pi_{\ell}\rangle]$,
processing $\Pi_\ell$. It is based on a sequence of vectors with $\vec X_0
= \langle \emptyset,\dots,\emptyset \rangle$ and $\vec X_{i+1} = F(\vec X_i)$
and determines the
interpretation of each $P_i$ ($1\leq i\leq m$) as the $i$-th component in the
fixed point $\vec X_\infty$ of this sequence (cf.\ the semantics of PDDL axiom
programs in Section \ref{sec:background}).

We use the same sequence of vectors to associate the ground atoms of these
predicates with different stages.
We say that a ground atom $P_i(\vec a)$ is \emph{derived in stage $l$} if $\vec
a$ occurs in the $i$-th component of $X_l$ but \emph{not} already in the $i$-th
component of $X_{l-1}$. If an atom is never derived, i.e.\ if it is false in
the interpretation defined by $\vec X_\infty$, then there is no such $l$.

Let $f$ be the stage where the fixed point for stratum $\Pi_\ell$ is reached,
i.e.\ $f$ is the least number for which $\vec X_f = \vec X_\infty$. For an
atom $P_i(\vec a)$, we write $\stage{\vec a}{i}$ to refer to the stage $l$ in
which the atom $P_i(\vec a)$ is derived, or to $f+1$ if there is no such $l$.

Before turning to the additional auxiliary predicates we use the stages to
define a number of auxiliary relations that will form the basis of these
predicates. Remember that $m$ is the number of
predicates affected by stratum $\Pi_\ell$. For $i,j\in \{1,\dots,m\}$, we
define the relation $\leij{i}{j}$ such that
\begin{equation}
  \vec a\leij{i}{j}\vec b\text{ iff }\stage{\vec a}{i}<\stage{\vec b}{j}.
\end{equation}
This means that $P_i(\vec a)$ is derived in a strictly earlier stage than
$P_j(\vec b)$, which possibly is not derived at all.

Analogously, we define relation $\leqij{i}{j}$ as
\begin{equation}
  \vec a\leqij{i}{j}\vec b\text{ iff }\stage{\vec a}{i}\leq\stage{\vec
b}{j}\text{ and } \stage{\vec a}{i}\leq f.
\end{equation}
This means that $P_i(\vec a)$ is derived in some stage and that this happens at
latest in the stage in which $P_j(\vec b)$ is derived (if the latter is derived
at all).

We also explicitly represent the complement relations $\nleij{i}{j}$ and
$\nleqij{i}{j}$, which are defined as
\begin{equation}
\vec a\nleij{i}{j}\vec b\text{ iff } \stage{\vec a}{i}\geq\stage{\vec b}{j}
\end{equation}
and as 
\begin{equation}
\vec a\nleqij{i}{j}\vec b\text{ iff }
\stage{\vec a}{i}>\stage{\vec b}{j}\text{ or }\stage{\vec a}{i}=f+1.
  \label{eq:nleqij}
\end{equation}

Lastly, we introduce the relation $\triij{i}{j}$ as
\begin{equation}
\vec a\triij{i}{j}\vec b\text{ iff }\stage{\vec a}{i} + 1 = \stage{\vec b}{j}.
\end{equation}

In the following, we write that $P_i(\vec a)$ is derived \emph{before},
\emph{strictly before}, and \emph{immediately before} $P_j(\vec b)$ if $\vec
a\leqij{i}{j}\vec b$, $\vec a\leij{i}{j}\vec b$, and $\vec a\triij{i}{j}\vec
b$, respectively.

In the proof of Theorem \ref{thm:stagelemma} we will show how to express these
relations by means of axioms.
For this purpose we
introduce corresponding predicate symbols with subscript ${\textup{ax}}$ (e.g.
$\leij{i}{j}_{\textup{ax}}$) to distinguish these new \emph{stage predicates}
from the corresponding \emph{stage relations}, which are induced by a specific
structure.
In the proof of Theorem \ref{thm:stagelemma}, we will present \emph{stage
axioms} that interpret these predicates for every basic structure as the
corresponding stage relation. We will also see that these axioms do not
introduce negative occurrences of any of these predicates.

Before we get to these details, we explain how the stage relations are relevant
for the overall compilation:
For each predicate $P_i$ affected by $\Pi_\ell$,
we can achieve the intended meaning of auxiliary predicate $\bar P_i$ as its
complement predicate ($\bar P_i(\vec x)$ being true iff $P_i(\vec x)$ is false)
by an additional \emph{complement axiom}
\begin{equation}
 \pddlaxiom{\bar{P_i}(\vec x)}{\vec x \nleqij{i}{i}_{\textup{ax}}\vec x}.
  \label{eq:compaxiom}
\end{equation}
The following theorem provides the theoretical basis for these complement axioms.
\begin{theorem}
  Let $\Pi$ be a stratum of an axiom program $\prog$, and
  $P_i$ be a predicate affected by $\Pi$. Let
  $\struct$ be a basic structure for $\prog$ and relation $\nleqij{i}{i}$ be
  defined as in equation \eqref{eq:nleqij}.

  For all $\vec a$ (of the arity of $P_i$) it holds that $\vec
  a \nleqij{i}{i}\vec a$ iff $\struct[\prog]\not\models P_i(\vec a)$.
\end{theorem}

\begin{proof}
  ``$\Rightarrow$'' Since $\vec a \nleqij{i}{i}\vec a$ iff $\stage{\vec
  a}{i}>\stage{\vec a}{i}\text{ or }\stage{\vec a}{i}=f+1$, and trivially
  $\stage{\vec a}{i}\not >\stage{\vec a}{i}$, it must hold that $\stage{\vec
  a}{i}=f+1$. By the definition of the stages, this means that $P_i(\vec a)$ is
  not in the fixed point for $\Pi$ and thus $\struct[\prog]\not\models P_i(\vec
  a)$.

  ``$\Leftarrow$'' If $\struct[\prog]\not\models P_i(\vec a)$ then $\stage{\vec
  a}{i} = f+1$, so $\vec a \nleqij{i}{i}\vec a$.
\end{proof}

\begin{algorithm}[t]
  \begin{algorithmic}[1]
  \Function{eliminate}{\genaxioms\ program $\prog=\langle \Pi_1,\dots,\Pi_n \rangle$}
      \For{$\Pi$ in $\Pi_1, \dots, \Pi_{n-1}$}
        \State$\textit{affected} := \{P \mid P$ is affected by $\Pi\}$
        \For{each $P_i, P_j\in \textit{affected}$}
          \State \parbox[t]{6cm}{add the axioms for $\leij{i}{j}_{\textup{ax}},
            \leqij{i}{j}_{\textup{ax}}, \nleij{i}{j}_{\textup{ax}},\nleqij{i}{j}_{\textup{ax}},$\\
            \phantom{add the axioms for }and $\triij{i}{j}_{\textup{ax}}$ to
            $\Pi$}
        \EndFor
        \For{each $P_i\in \textit{affected}$}
          \State add axiom $\pddlaxiom{\bar{P_i}(\vec x)}{\vec x
          \nleqij{i}{i}_{\textup{ax}}\vec x}$ to $\Pi$
        \EndFor
      \EndFor
      \For{each derived $P_i$ occurring negatively in \prog}
        \State \parbox[t]{6cm}{replace every negative occurrence $P_i(\vec x)$
        in $\prog$ with $\lnot\bar{P_i}(\vec x)$}
      \EndFor
      \State \Return $\prog$
  \EndFunction
\end{algorithmic}
  \caption{Elimination procedure}
\label{alg:eliminationprocedure}
\end{algorithm}

Based on the complement axioms we can transform a given axiom program $\prog
= \langle \Pi_1, \dots, \Pi_n \rangle$ as shown in Algorithm
\ref{alg:eliminationprocedure}.
In a first pass, we add to each stratum the stage axioms, as defined in the
proof of Theorem \ref{thm:stagelemma}, and the complement axioms, as in Equation
\eqref{eq:compaxiom}. We can skip the last stratum because its affected
predicates cannot occur negatively in a stratified program.
In a second pass, we can then eliminate all negative occurrences of derived
predicates, replacing them with their negated complement predicates.

We have now presented all components of the compilation except for the stage
axioms. We conclude by showing that the stage relations can indeed be expressed
by axioms.

\begin{theorem}
  The relations $\leij{i}{j}$, $\leqij{i}{j}$, $\nleij{i}{j}$, $\nleqij{i}{j}$
  and $\triij{i}{j}$ can be defined by an \semposaxioms\ program.
  \label{thm:stagelemma}
\end{theorem}
  
In the following proof, we will use subformulas of the form $\varphi(\vec
x)[\leqi{j}{\vec y}]$.
These mean that in $\varphi(\vec x)$ every occurrence of an atom $P_k(\vec z)$
with $k\in\{1,\dots,m\}$ is replaced by $\vec z\leqij{k}{j}_{\textup{ax}}\vec
y$. Likewise for $\varphi(\vec x)[\lei{j}{\vec y}]$.

For example, for $\varphi(x,x') = \exists x'' (P_1(x,x'') \land
P_2(x'',x'))$ where $P_1,P_2$ are derived on stratum $\Pi_\ell$, the formula
$\varphi(x, x')[\leqi{2}(y,y')]$ is $\exists x''
((x,x'')\leqij{1}{2}_{\textup{ax}}(y,y')
\land (x'',x')\leqij{2}{2}_{\textup{ax}}(y,y'))$.
It corresponds to $\varphi$, where in the evaluation we may only
use the atoms derived before $P_2(y,y')$.

Similarly, we use formulas of the form $\varphi(\vec x)[\lnot\nleqi{j}{\vec y}]$
that replace every occurrence of 
$P_k(\vec z)$ by $\lnot \vec z \nleqij{k}{j}_{\textup{ax}}\vec y$.
Likewise for $\varphi(\vec x)[\lnot\nlei{j}{\vec y}]$. These will be used
if we negate the formulas, so that overall all occurrences are positive
again. The last kind of subformula is $\varphi(\vec x)[\bot]$, replacing all
occurrences of any $P_k(\vec z)$ by $\bot$ (\emph{false}). It is true, if
$\varphi$ is already true during the computation of the first stage.

\begin{proof}
  Let $P_1, \dots, P_m$ be the predicates affected by a stratum $\Pi_\ell$.
  We assume w.l.o.g.\ that all axioms in stratum $\Pi_\ell$ use distinct
  variables and that $\Pi_\ell$ contains for each $P_i$ only a single axiom
  that affects $P_i$.  Otherwise we can combine the bodies of such axioms in a
  disjunction, renaming the variables accordingly. We refer to the
  body of the axiom affecting $P_i$ as $\varphi_i(\vec x)$.

  For $i,j\in\{1,\dots,m\}$, we use the following axioms (explained below):
  \begin{align}
      \pddlaxiom{\vec x\leij{i}{j}_{\textup{ax}} \vec
      y}{&\bigvee\nolimits_{k=1}^{m}\exists\vec z (\vec
      x\leqij{i}{k}_{\textup{ax}}\vec z \land \vec z
      \triij{k}{j}_{\textup{ax}}\vec y)} \label{eq:leijaxiom}\\
      \pddlaxiom{\vec x\leqij{i}{j}_{\textup{ax}} \vec y}{&\varphi_i(\vec
      x)[\lei{j}{\vec y}]} \label{eq:leqijaxiom}\\
      \pddlaxiom{\vec x\nleij{i}{j}_{\textup{ax}}\vec
      y}{&\varphi_j(\vec y)[\bot]\lor{}\label{eq:nleijaxiom}\\
      &\big(\bigvee\nolimits_{k=1}^{m}
      \exists\vec z (\vec x\nleqij{i}{k}_{\textup{ax}}\vec z \land \vec
      z \triij{k}{j}_{\textup{ax}}\vec y)\big)
      \lor{}\nonumber\\
      &\big(\bigwedge\nolimits_{k=1}^{m}\forall \vec z \lnot\varphi_k(\vec
      z)[\bot]\big)}\nonumber\\
      \pddlaxiom{\vec x\nleqij{i}{j}_{\textup{ax}}\vec y}{
        &\lnot\varphi_i(\vec x)[\lnot\nlei{j}{\vec y}]}\label{eq:nleqijaxiom}\\
      \pddlaxiom{\vec x\triij{i}{j}_{\textup{ax}}\vec y}{
       &\varphi_i(\vec x)[\lei{i}{\vec x}]\land
      \lnot\varphi_j(\vec y)[\lnot\nlei{i}{\vec x}]\land{}\label{eq:triijaxiom}\\
      &\big(\varphi_j(\vec y)[\leqi{i}{\vec x}]\lor{}\nonumber\\
      &\bigwedge\nolimits_{k=1}^{m}\forall \vec z(\lnot\varphi_k(\vec
        z)[\lnot\nleqi{i}{\vec x}]\lor\varphi_k(\vec z)[\lei{i}{\vec x}])\big)
      }\nonumber
  \end{align}

Note that these axioms have no negative occurrences of a stage predicate or
of a predicate affected by $\Pi_\ell$.

Eq.\ \eqref{eq:leijaxiom} expresses that $P_i(\vec{x})$ is derived strictly
before $P_j(\vec{y})$ if it is derived before some $P_k(\vec{z})$, which is in
turn derived immediately before $P_j(\vec{y})$.

Eq.\ \eqref{eq:leqijaxiom} states that $P_i(\vec{x})$ is derived before
$P_j(\vec{y})$ because $P_i(\vec{x})$ can already be derived using only atoms that are
derived strictly before $P_j(\vec{y})$.

In its three disjuncts, eq.\ \eqref{eq:nleijaxiom} lists three possibilities
why $P_i(\vec{x})$ is not derived strictly before $P_j(\vec{y})$: (a)
$P_j(\vec{y})$ is already derived in stage 1, (b) there is some $P_j(\vec{z})$
derived immediately before $P_j(\vec{y})$ and $P_i(\vec{x})$ is not derived
before this $P_k(\vec{z})$, so $P_i(\vec{x})$ is not derived strictly before
$P_j(\vec{y})$, or (c) nothing can be derived at all, so both atoms are in the
same stage $f+1$.

Eq.\ \eqref{eq:nleqijaxiom} states that $P_i(\vec{x})$ is not derived before
$P_j(\vec{y})$ because it cannot be derived using only the atoms derived
strictly before $P_j(\vec{y})$ (expressing $\prec$ as negated $\nprec$ to
avoid a negated occurrence of $\prec$ in the overall negated formula).

In its conjuncts, eq.\ \eqref{eq:triijaxiom} lists three
requirements for $P_i(\vec{x})$ being derived immediately before
$P_j(\vec{y})$: (a) $P_i(\vec{x})$ can be derived from the atoms derived
strictly before $P_i(\vec{x})$, implying that it is true in the fixed point, (b)
$P_j(\vec{y})$ cannot be derived from the atoms derived strictly before
$P_i(\vec{x})$, implying that is not derived at the same stage as $P_i(\vec x)$
(or earlier), and (c) $P_j(\vec{y})$ can be derived from the atoms derived
before $P_i(\vec{x})$, or $P_i(\vec{x})$ was derived in the stage that reached
the fixed point (and $P_j(\vec{y})$ is false in the fixed point). The last
property is expressed by the requirement that all $P_k(\vec{z})$ that can be
derived from the atoms derived before $P_i(\vec{x})$ can also be derived from
the atoms derived \emph{strictly} before $P_i(\vec{x})$.

We will first prove that these relations establish a fixed point for these
axioms, and then separately confirm that we can actually derive the corresponding
interpretation from these axioms. 

  Let $\struct'$ interpret the five predicates as the corresponding relations
  for some basic structure $\struct$.
  To show that $\struct'$ corresponds to a fixed point of
  $\Pi_{\textit{stage}}$, we can treat each axiom independently.

  \begin{itemize}
    \item Axiom \eqref{eq:leijaxiom}: If for some $k$ there is $\vec z$ so
      that $(\vec x\leqij{i}{k}_{\textup{ax}}\vec z \land \vec z \triij{k}{j}_{\textup{ax}}\vec y)$
      is true under $\struct'$ then $\stage{\vec x}{i}\leq\stage{\vec
      z}{k}$, $\stage{\vec x}{i}\leq f$ and $\stage{\vec z}{k}
      + 1 = \stage{\vec y}{j}$ (because we interpret
      $\vec x\leqij{i}{k}_{\textup{ax}}\vec z$ as $\vec x\leqij{i}{k}\vec z$ and
      $\vec z \triij{k}{j}_{\textup{ax}}\vec y$ as $\vec z \triij{k}{j}\vec y$).
      We can conclude that $\stage{\vec
      x}{i}<\stage{\vec y}{j}$ and $\vec x\leij{i}{j} \vec y$, so
      $\struct'\models\vec x\leij{i}{j}_{\textup{ax}} \vec y$.
    \item Axiom \eqref{eq:leqijaxiom}: If $\varphi_i(\vec
      x)[\lei{j}{\vec y}]$ holds then the atoms derived in a stage
      $l<\stage{\vec y}{j}$ are sufficient to derive $P_i(\vec x)$. We can
      conclude that $\stage{\vec x}{i}\leq\stage{\vec y}{j}$. Moreover, since
      a strictly earlier stage implies that these predicates are derivable, we
      can actually derive $P_i(\vec x)$, so $\stage{\vec x}{i}\leq f$. Together
      we have $\vec x\leqij{i}{j} \vec y$.
    \item  Axiom \eqref{eq:nleijaxiom}: 
      If $\varphi_j(\vec y)[\bot]$ holds, then $P_j(\vec y)$ can be derived in
      stage $1$, and we trivially have $\vec x\nleij{i}{j}\vec y$.

      If $\struct'$ satisfies $\bigvee_{k=1}^{m} \exists\vec
      z (\vec x\nleqij{i}{k}_{\textup{ax}}\vec z \land \vec z \triij{k}{j}_{\textup{ax}}\vec y)$,
      there is a $k$ such that $\stage{\vec z}{k} + 1 = \stage{\vec y}{j}$ and
      $\stage{\vec x}{i}>\stage{\vec z}{k}$ or $\stage{\vec x}{i}=f+1$. Thus
      $\stage{\vec x}{i}\geq\stage{\vec y}{j}$ and $\vec x\nleij{i}{j}\vec
      y$.
      
      If $\struct'$ satisfies the third disjunct in the body, nothing can be
      derived in the entire stratum, so $f=0$ and $\stage{\vec
      x}{i}=\stage{\vec y}{j}=1$, again implying $\vec x\nleij{i}{j}\vec y$.
    \item Axiom \eqref{eq:nleqijaxiom}: If $\varphi_i(\vec x)[\lnot\nlei{j}{\vec y}]$
      does not hold, $P_i(\vec x)$ cannot be derived from the axioms that are
      derived at all stages $l<\stage{\vec y}{j}$, so $\stage{\vec
      x}{i}>\stage{\vec y}{j}$ if $P_j(\vec y)$ is derivable, otherwise $\stage{\vec
      x}{i}=\stage{\vec y}{j} = f + 1$. We conclude that $\vec
      x\nleqij{i}{j}\vec y$.
    \item Axiom \eqref{eq:triijaxiom}: From the first conjunct, we get 
      $\stage{\vec x}{i} \leq f$, from the second one that $\stage{\vec y}{j}
      > \stage{\vec x}{i}$. The first disjunct in the last conjunct implies $\stage{\vec y}{j}
      \leq \stage{\vec x}{i} + 1$, the second one that
      every atom $P_k(\vec z)$ cannot be derived up to stage $\stage{\vec
      x}{i}+1$ or has already been derived at stage $\stage{\vec x}{i}$ or before. So
      the conjunction (from $k=1$ to $m$) expresses that the fixed point has been reached at
      stage $\stage{\vec x}{i}$. Thus, the first part of the disjunction covers the case
      where $P_j(\vec y)$ is derivable and the second part the one where it is
      not. In both cases, we get that $\stage{\vec y}{j} = \stage{\vec x}{i}
      + 1$ and thus $\vec x\triij{i}{j}\vec y$.
  \end{itemize}

  To establish that $\struct'$ corresponds to the \emph{least} fixed point of
  $\Pi_{\textit{stages}}$, we show for all relations
  $R\in\{\leqij{i}{j},\leij{i}{j},$ $\nleqij{i}{j},\nleij{i}{j},\triij{i}{j}\mid
  i,j\in\{1,\dots,m\}\}:=\mathcal R$ that if $\vec a R\vec b$ holds (defined
  relative to a specific basic structure $\struct$) then $(\vec a, \vec b)$ is
  in the interpretation of $R_{\textup{ax}}$ as computed by the extension of
  $\struct$ with axioms $\Pi_\textit{stage}$, i.e.\ it can be derived from the
  axioms. We prove this by induction over $\stage{\vec b}{j}$, starting with
  the induction basis $\stage{\vec b}{j} = 1$:
    
  \begin{itemize}
    \item Case $\triij{i}{j}$: If $\stage{\vec b}{j} = 1$ there is no $\vec a$
      with $\vec a\triij{i}{j}\vec b$, so the statement is trivially true.
    \item Case $\leqij{i}{j}$: If $\vec a\leqij{i}{j}\vec b$ then $\stage{\vec
      a}{i}\leq 1$ and $\stage{\vec a}{i}\leq f$. If $f=0$, nothing can be
      derived from the stratum and there is no such $\vec a$.
      Otherwise $P_i(\vec a)$ is derived in the first stage and $\varphi_i(\vec
      a/\vec x)[\bot]$ is true, so also $\varphi_i(\vec a/\vec x)[\lei{j}{\vec b}]$
      is true and we can derive $\vec a\leqij{i}{j}_{\textup{ax}}\vec b$.
    \item Case $\leij{i}{j}$: If $\vec a\leij{i}{j}\vec b$ then $\stage{\vec
      a}{i}< 1$. There is no such $\vec a$, so the statement is trivially true.
    \item Case $\nleij{i}{j}$: If $\stage{\vec b}{j} \leq f$, then 
      $\varphi_j(\vec b/\vec y)[\bot]$ holds and we can derive $\vec
      a\nleij{i}{j}_{\textup{ax}}\vec b$.
      If $\stage{\vec b}{j} > f$, nothing can be derived from the
      stratum and we can use the third disjunct of the body of
      \eqref{eq:nleijaxiom} to derive $\vec a\nleij{i}{j}_{\textup{ax}}\vec b$.
    \item Case $\nleqij{i}{j}$: If $\vec a\nleqij{i}{j}\vec b$ then
      $\stage{\vec a}{i}>\stage{\vec b}{j}$ or $\stage{\vec a}{i}=f+1$,
      implying $\stage{\vec a}{i}>1$.
      So $P_i(\vec a)$ cannot be derived in stage $1$ and
      $\varphi_i(\vec a/\vec x)[\bot]$ must be
      false. As $\stage{\vec b}{j} = 1$, this implies that
      $\varphi_i(\vec a/\vec x)[\lnot\nlei{j}{\vec b}]$ is false and we can
      derive $\vec a\nleqij{i}{j}_{\textup{ax}}\vec b$.
  \end{itemize}

  For the induction hypothesis, suppose that it holds for all relations
  $R\in\mathcal R$ that if $\vec a R\vec b$ and $\stage{\vec b}{j}\leq l$ then
  $\vec a R_{\textup{ax}}\vec b$ can be derived from $\Pi_{\textit{stages}}$.
\medskip
  
  Inductive step: $\stage{\vec b}{j} = l+1$
    
  \begin{itemize}
    \item Case $\triij{i}{j}$: Suppose that $\vec a \triij{i}{j}\vec b$, so
      $\stage{\vec a}{i} = l$.
      In the following, we explain for each of the three conjuncts in the body
      of axiom \eqref{eq:triijaxiom} (instantiating $\vec x$ with $\vec a$ and
      $\vec y$ with $\vec b$) why it is true if for all five stage comparison
      predicates $R$, $\vec cR_{\textup{ax}}\vec a$ has been derived for all $\vec c$
      with $\vec cR\vec a$. Since $\stage{\vec a}{i} = l$, this is the case
      by the induction hypothesis.

      For the first conjunct, we use $\stage{\vec a}{i} = l$:
      $P_i(\vec a)$ can be derived by the stratum only using atoms derived up to
      stage $l-1$.

      For the second conjunct, observe that $P_j(\vec b)$ cannot be derived by
      the stratum only using atoms derived up to stage $l-1$ (otherwise its
      stage was $\leq l$), so the subformula in the negation is false.
      
      For the last conjunct, if $l+1\leq f$ (i.e.\ $P_j(\vec b)$ can be derived
      by the stratum), we focus on its first disjunct and use the same argument as
      for the first conjunct. If $l+1 > f$, the fixed point for the stratum
      is reached at stage $l$. In this case, it holds for all atoms
      with predicates from this stratum that they can already be derived given
      the atoms derived up to stage $l-1$ or they cannot be derived given in
      addition the atoms from stage $l$. This is expressed by
      $\bigwedge\nolimits_{k=1}^{m}\forall \vec z(\lnot\varphi_k(\vec
      z)[\lnot\nleqi{i}{\vec a}]\lor\varphi_k(\vec z)[\lei{i}{\vec a}])$ (the
      last part of axiom \eqref{eq:triijaxiom} where $\vec x$ is replaced by
      $\vec a$).

    \item Case $\nleij{i}{j}$. Suppose that $\vec a\nleij{i}{j}\vec b$, i.e.\
      $\stage{\vec a}{i}\geq\stage{\vec b}{j}$. If the first or third disjunct
      in the body of \eqref{eq:nleijaxiom} is true then the claim follows trivially.
      Therefore, suppose that the first and third disjunct are false. We show
      that then the second disjunct is true.

      As the first and third one are false, we have $\stage{\vec b}{j}>1$ and the
      fixed point of stratum $\Pi_\ell$ is not empty. Thus there is some
      $P_k(\vec c)$ derived in stage $l$, i.e.\ $\vec c \triij{k}{j}\vec b$.
      By the induction hypothesis, $\vec a \nleqij{i}{k}_{\textup{ax}}\vec c$ can be
      derived from $\Pi_\textit{stage}$, and we can use the previous case for $\vec
      c\triij{k}{j}_{\textup{ax}}\vec b$ (where the argument only relies on stage
      predicates for smaller stages).
    \item Case $\leij{i}{j}$. Suppose that $\vec a\leij{i}{j}\vec b$, i.e.\
      $\stage{\vec a}{i}<\stage{\vec b}{j}$. Then $l+1=\stage{\vec b}{j}>1$ and
      there is some atom $P_k(\vec c)$ derived in stage $l$. For this $k$ and
      $\vec c$, we have $\vec a \leqij{i}{k}\vec c$ and $\vec
      c\triij{k}{j}\vec b$. As in the previous case, we use the induction
      hypothesis for $\vec a \leqij{i}{k}_{\textup{ax}}\vec c$ and the argument from
      case $\nleij{i}{j}$ for $\vec c \triij{k}{j}_{\textup{ax}}\vec b$ to finish the
      proof for this case.
    \item Case $\leqij{i}{j}$. Suppose that $\vec a\leqij{i}{j}\vec b$, i.e.\
      $\stage{\vec a}{i}\leq\stage{\vec b}{j}$ and $\stage{\vec a}{i}\leq f$. 
      Then $P_i(\vec a)$ can be derived based on the atoms derived up to stage
      $l$, which is expressed by the body of \eqref{eq:leqijaxiom}. The
      derivability of the stage predicates was established by case
      $\leij{i}{j}$.
    \item Case $\nleqij{i}{j}$. Suppose that $\vec a\nleqij{i}{j}\vec b$, i.e.\ 
      $\stage{\vec a}{i}>\stage{\vec b}{j}$ or $\stage{\vec a}{i}= f+1$. This
      means the atoms derived up to stage $l-1$ are not sufficient to
      derive $P_i(\vec a)$, i.e.\ to make $\varphi_i$ true. This is expressed by
      the body of \eqref{eq:nleqijaxiom}. The
      derivability of the relevant stage predicates was established by case
      $\nleij{i}{j}$.
  \end{itemize}
\end{proof}

The theoretical result in Section \ref{sec:equivalence} implied that negative
occurrences can be eliminated, but it did not explain how this can be achieved.
Addressing this gap in understanding was our primary motivation for presenting
this transformation. For this reason, we prioritized clarity over minimizing
the size of the resulting axiom program.

There is a number of possible straight-forward improvements: we only introduced
the complement predicates for instructional reasons, but could easily omit them
and the complement axioms by using directly the body of these axioms in the
replacement of negative occurrences; instead of introducing the stage axioms
for \emph{all} predicates in each stratum, we could analyze which of these are
necessary based on the predicates with actual negative occurrences; last, many
stage axioms contain the same subexpressions (e.g.\ the large conjunction in
Eq.\ \eqref{eq:triijaxiom} does not depend on $j$). These could be represented
only once by means of additional axioms.

\section{Conclusion and Future Work}

We considered three axiom formalism used in automated planning. The most
general one, $\genaxioms$, allows arbitrary first-order formulas in axiom bodies
as long as the axioms are stratifiable. The other two impose different
syntactic restrictions: $\semposaxioms$ forbids negative occurrences of derived
predicates in axiom bodies or, equivalently, only permits a single stratum.
Stratified Datalog restricts bodies to existentially quantified conjunctions
of literals.

As a measure for expressiveness, we compared which queries these
formalisms can express. We showed that $\genaxioms$ and $\semposaxioms$
are equally expressive, matching the expressive power of least fixed point
logic, and thus strictly exceeding the expressiveness of stratified Datalog.
This is the strongest possible negative result: Regardless of what time and
space we can use in the compilation, some $\genaxioms$ and $\semposaxioms$
programs cannot be expressed in stratified Datalog.

Unlike \inlinecite{thiebaux-et-al-aij2005}, we examined axiom languages
independently of the other components of the planning task and therefore did
not consider compilations that shift part of the axiom evaluation into
action applications. Using the corresponding compilation-scheme framework 
\cite{nebel-jair2000,thiebaux-et-al-aij2005}, it may still be possible to
compile tasks with $\genaxioms$ axioms into tasks with stratified Datalog
axioms if a non-constant increase of the plan length is allowed.

However, our positive  result on the equivalence of $\genaxioms$ and $\semposaxioms$
also applies in this setting: the elimination procedure from Section
\ref{sec:elimination} directly gives rise to a polynomial-time compilation
scheme that preserves plan size exactly.

Our primary motivation for the elimination procedure was to obtain a deeper
theoretical understanding of the equivalence of the two axiom formalisms.
In future work, we will also evaluate its practical usefulness empirically.
A potential obstacle is the induced blow-up: Although it is only polynomial on
the lifted level, it may be prohibitive for planners that ground the axioms.
The arity of derived predicates is a particularly critical factor, as the
maximum arity will inevitably double for many axiom programs.

Moreover, the compilation introduces auxiliary predicates that are relevant
only during the axiom evaluation but not for action applicability or for
satisfying the goal.  Such atoms increase the size of the state representation
in the planning task. In forward state-space search, however, storing such
atoms can be avoided, a general optimization that should be especially
beneficial when using the compilation.

\section*{Acknowledgements}
We have received funding for this work from the Swiss National Science
Foundation (SNSF) as part of the project ``Lifted and Generalized
Representations for Classical Planning'' (LGR-Plan).

\end{document}